\pdfoutput=1
\documentclass[11pt]{article}

\usepackage{ACL2023}
\usepackage{times}
\usepackage{latexsym}
\usepackage[T1]{fontenc}
\usepackage[utf8]{inputenc}
\usepackage{microtype}
\usepackage{inconsolata}
\usepackage{float}
\usepackage{graphicx}
\usepackage{tikz}
\usepackage{lineno}
\usepackage{amsmath}
\usepackage{multirow}
\usepackage{booktabs}
\usepackage{array}
\usetikzlibrary{shapes, arrows.meta, positioning}
\usepackage{amssymb}
\usepackage{xcolor}
\usepackage[dvipsnames]{xcolor}
\usepackage[table]{xcolor}

\definecolor{ourrow}{RGB}{245,245,255} 
\usepackage[shortlabels]{enumitem}

\usepackage{dblfloatfix}  

\title{EL-MIA: Quantifying Membership Inference Risks of Sensitive Entities in LLMs}

\author{
Ali Satvaty\textsuperscript{1} \quad 
Suzan Verberne\textsuperscript{2} \quad 
Fatih Turkmen\textsuperscript{1} \\
\textsuperscript{1}University of Groningen \quad
\textsuperscript{2}Leiden University \\
\textit{Correspondence:} \texttt{a.satvaty@rug.nl}
}
\setcitestyle{round}            
\begin{document}
\maketitle

\begin{abstract}
Membership inference attacks (MIA) aim to infer whether a particular data point is part of the training dataset of a model.
In this paper, we propose a new task in the context of LLM privacy: entity-level discovery of membership risk focused on sensitive information (PII, credit card numbers, etc).
Existing methods for MIA can detect the presence of entire prompts or documents in the LLM training data, but they fail to capture risks at a finer granularity.
We propose the ``EL-MIA'' framework for auditing entity-level membership risks in LLMs. We construct a benchmark dataset for the evaluation of MIA methods on this task. Using this benchmark, we conduct a systematic comparison of existing MIA techniques as well as two newly proposed methods.
We provide a comprehensive analysis of the results, trying to explain the relation of the entity level MIA susceptability with the model scale, training epochs, and other surface level factors.
Our findings reveal that existing MIA methods are limited when it comes to entity-level membership inference of the sensitive attributes, while this susceptibility can be outlined with relatively straightforward methods, highlighting the need for stronger adversaries to stress test the provided threat model.
 \end{abstract}

\section{Introduction}





LLMs have demonstrated remarkable capabilities across a wide range of natural language processing tasks, but their deployment raises growing privacy concerns ~\cite{zhao2025surveylargelanguagemodels, llm_evaluation, yan2024protectingdataprivacylarge, dong-etal-2024-attacks}. A key risk stems from the memorization of personally identifiable information (PII) present in training data \cite{carlini2021extracting, lukas2023analyzing, zhou2024quantifying, satvaty2025undesirable}. Prior studies have shown that LLMs can regurgitate sensitive data verbatim or in paraphrased form, enabling membership inference attacks (MIAs) that reveal whether a specific record was used during training. Such leakage has serious implications for user privacy, regulatory compliance (e.g., GDPR, CCPA), and the safe use of LLMs in real-world applications.


While existing MIA methods have made significant progress in detecting whether entire documents were present in the training set, it is unclear to what extent they can capture risks at a finer granularity, specifically, entity-level membership risk for individual sensitive attributes such as names, dates of birth, addresses or phone numbers \cite{mia_sok}. This is of practical relevance since attackers often seek to recover or confirm the presence of particular sensitive fields rather than entire text samples. 

In this work, we introduce a framework for auditing entity-level membership risks in LLMs. Our approach targets the vulnerability of specific sensitive entities in LLM training, enabling more precise privacy risk assessments and supporting the development of stronger defenses.
\begin{figure}[t]
    \centering
    \fbox{\includegraphics[width=.46\textwidth]{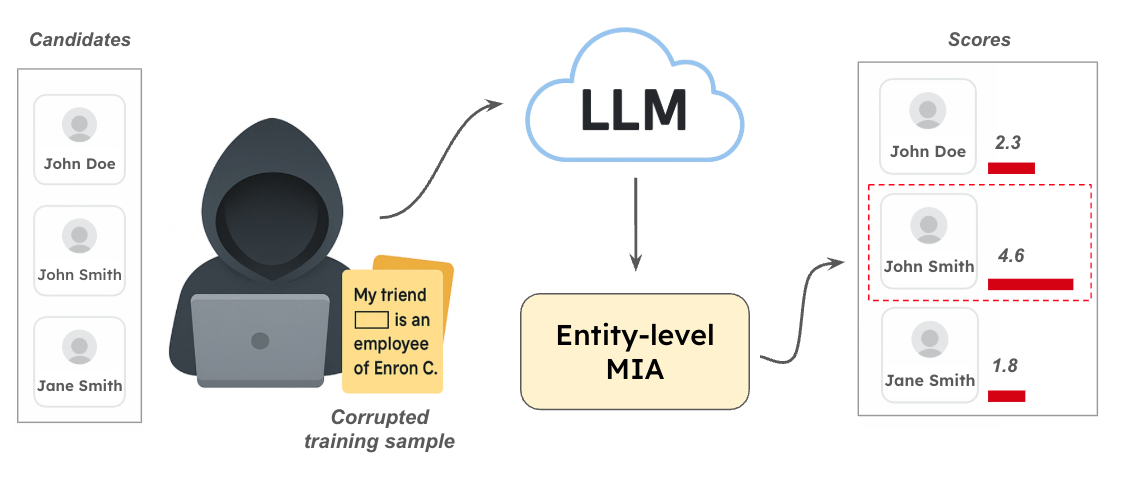}}
    \caption{Entity membership inference threat model}
    \label{fig:overall}
\end{figure}
We construct a benchmark dataset for evaluating slot-level PII MIA, based on the existing AI4Privacy dataset \cite{ai4privacy_2023}. Using the newly created benchmark, we conduct a systematic comparison of existing MIA techniques, assessing their ability to detect membership of individual sensitive fields rather than long input texts. We also propose two novel methods specifically aimed at entity-level MIA attacks. Our approach reduces the noise of the token-loss signal by normalizing across a reference set and omitting the effect of prefix tokens' loss. By empirical evaluation we compare the effectiveness of existing as well as new methods. 
We further perform extensive analyses regarding token length, attribute types, number of links,  model size, and training dynamics.

In summary, our contributions are threefold: (1) we propose entity-level MIA tailored to the sensitive entities as a new threat model in the context of privacy in LLMs; (2) we present a new dataset, built on top of the AI4Privacy benchmark, to evaluate methods on the newly proposed threat model;\footnote{We release the code for creating the dataset from the original AI4Privacy data as well as our methods: \url{https://github.com/alistvt/el-mia-data}} (3) we evaluate the effectiveness of existing and new methods for entity-level MIA and explain the dynamics of the inference susceptibility of the samples based on various cues.

\begin{table}[ht]
\centering
\scriptsize
\setlength{\tabcolsep}{6pt}
\renewcommand{\arraystretch}{1.2}
\begin{tabular}{p{1.6cm} p{2.5cm} p{2.3cm}}
\toprule
\textbf{Membership unit} & \textbf{Work(s)} & \textbf{Granularity} \\
\midrule
\multirow{4}{*}{\textbf{Sequence-level}}
& \citet{domembershipwork} & $\sim$n-grams \\
& \citet{carlini2021extracting, min_k_prob} & 32–256 tokens \\
& \citet{kaneko2024sampling, liu2024probing, neighbourcomparison2023}, \citet{neighbourcomparison2023} & Variable lengths \\
\midrule
\multirow{2}{*}{\textbf{Document-level}}
& \citet{meeus2024did, min_k_prob, speakmemory2023} & Full articles, books \\
\midrule
\textbf{Dataset-level}
& \citet{maini2024llm, hayes2025strongmembershipinferenceattacks} & Whole benchmarks/datasets \\
\midrule
\textbf{Entity-level}
& \textbf{This work} & Entities + Templates \\
\bottomrule
\end{tabular}
\caption{LLM membership inference attacks by prior work, grouped by the unit of membership (sequence, document, dataset, entity). Our proposed model targets \textbf{entity-level membership}, probing whether a specific sensitive value (e.g., a name or phone number) appeared in training, given the respective context.}
\label{tab:grouped_mia}
\end{table}

\section{Background and Related Work}
Membership inference was first studied in genomics, where researchers asked whether an individual’s data contributed to an aggregate dataset \cite{homer2008resolving}. The idea later migrated to machine learning, with the seminal attack against deep neural networks showing that models can reveal whether a specific example appeared in training \cite{shokri2017membership}. Whereas traditional MIAs in machine learning typically operate on a clearly defined unit such as a single record or image, the notion of membership is more ambiguous in the text domain. 

\subsection{Granularity of text}
For text data, most existing work has defined the unit of membership inference as sequences of fixed (often truncated) length. For example, \citet{carlini2021extracting} employ sequences of 256 tokens, whereas \citet{min_k_prob} and subsequent studies \cite{kaneko2024sampling, liu2024probing, mozaffari2024semantic, wang2025con, xie2024recall, ye2024data, zhang2024adaptive, zhang2024min} rely on random excerpts ranging between 32 and 256 tokens. When sequences are randomly injected, prior work has considered lengths of up to 80 tokens \cite{wei202proving} or 100 tokens \cite{copyright2024trap}.

In contrast, some research aims to detect membership at a much larger scale, such as an entire book \cite{meeus2024did, min_k_prob, speakmemory2023} or even a whole dataset \cite{maini2024llm, hayes2025strongmembershipinferenceattacks}. In copyright-related settings, for instance, the objective is to determine whether the full content of a source (e.g., a news article or book) was used, rather than only specific excerpts. This has led prior work \cite{zhang2024adaptive, meeus2024did, decop2024} to pursue document- or dataset-level membership inference, where signals are aggregated across broader text units. This progression from fixed-length sequences to document- and dataset-level inference is further summarized in Table~\ref{tab:grouped_mia}.

Yet, to the best of our knowledge, no prior work has systematically investigated finer-grained units, such as individual entities \cite{mia_sok}. Our work aims to close this gap by introducing a benchmark for entity-level membership inference, with an emphasis on sensitive entities, as these pose the most significant privacy and security risks in practical LLM deployments. 




\subsection{Entity-level attacks on LLMs}
Previous work, while not directly formulating entity-level MIAs, has examined memorization and leakage of specific entities in LLMs. Early studies introduce canary insertion and the exposure metric to quantify sequence-level memorization \citep{carlini2021extracting, carlini2019secret}, later adapted to NLU classifiers and auditing frameworks \citep{parikh-etal-2022-canary}. Similar ideas were applied to domain-specific models, e.g., probing clinical BERT for sensitive patient-condition associations and testing pseudonymization \citep{vakili2021clinical, vakili2023using}.

More recently, researchers have begun to analyze entity-level memorization explicitly, showing that LLMs can reconstruct missing named entities from partial prompts \citep{zhou2024quantifying, kim2023propile}. At the same time, stronger auditing methods \citep{panda2025privacy, lukas2023analyzing, hayes-etal-2025-measuring} and practical PII extraction attacks \citep{cheng2025effective} demonstrate that identifiers can be surfaced.

While these approaches demonstrate valuable insights into memorization and leakage risks of sensitive entities, they fall short in comprehensively evaluating the leakage across the diverse range of sensitive entity types (e.g., names, phone numbers, account identifiers). Our work aims to fill this gap by introducing the first entity-level membership inference attack on LLMs.






\section{Threat Model}
In this section, we present our threat model and discuss how it differs from the existing threat models. 
\vspace{-.6cm}
\subsection{Attacker's Objective}
The attacker's objective is to determine whether a \textbf{specific entity} (e.g., a person's full name) was part of the training data of the LLM, \textbf{given a partial or redacted sentence} (e.g., "<full name> teaches at the University of New York.") and a candidate entity value (e.g., John Doe).






Formally, let $\mathcal{V}$ denote the set of candidate entity values and let $\mathcal{X}$ denote the space of textual samples (strings or token sequences). A \textbf{template} is an operator $T:\mathcal{V}\to\mathcal{X}$ that maps a value $v\in\mathcal{V}$ to a concrete sample $s=T(v)\in\mathcal{X}$. Let $D_{\mathrm{train}}\subseteq\mathcal{X}$ be the (private) training corpus of the model. The \textbf{entity-level membership indicator} for a candidate value $v$ under template $T$ is the binary variable

$$
M(v,T)=
\begin{cases}
1 & \text{if } T(v)\in D_{\mathrm{train}},\\[4pt]
0 & \text{if } T(v)\notin D_{\mathrm{train}}.
\end{cases}
$$

An attacker $\mathcal{A}$ observes some side information about training samples (the template $T$), and possibly additional auxiliary data or model outputs (candidate $v$), and outputs either: \textbf{(1)} a score $p\in[0,1]$ estimating $\Pr\big(M(v,T)=1\big)$ or \textbf{(2)} a binary decision $\widehat{M}\in\{0,1\}$ (obtained by thresholding $p$). The attacker's goal is to accurately decide on $M(v,T)$.

\subsection{Attacker Capabilities}
We make the following assumptions about the attacker:

\textbf{Model access.} The adversary has black-box or white-box access to the trained model; in our experiments, we assume token-level log-probabilities are queryable.
    
\textbf{Format knowledge.} The adversary knows the data distribution or template structure used to generate texts (e.g., sentence forms that contain PII slots).
    
\textbf{Candidate set.} The adversary holds a set of plausible sensitive entity values $V$ (e.g., names, phone numbers), some of which may appear in training paired with the known templates.
    
\textbf{Partial sample access.} The adversary may observe a \emph{redacted} or \emph{canonicalized} version of a training sample, i.e., the template with the sensitive entity field masked.


\subsection{Applicability to Real-World Scenarios}


The proposed threat model is designed to capture realistic privacy risks that arise in operational LLM deployments. In many cases, an attacker does not have access to complete unredacted training data, but may possess partial information such as a document template or prompt structure. For example, a redacted dataset or a publicly available template (e.g., a partially anonymized medical record, legal case file, or chat log) may reveal the general format of the text while concealing the actual PII values. The attacker’s objective is then to recover the hidden fields.

In such situations, 
an adversary who knows the overall data format -- such as the typical sequence of fields in an electronic health record, a customer support transcript, or a social media post -- can construct plausible candidate values for the missing PII slots. 
In multi-entity contexts, an attacker may already know $n-1$ sensitive entities appearing in related records and wish to infer the $n$th entity based on the correlations captured during training. For example, in a partially known patient cohort or group chat, if all but one participant’s identity is known, the model can help the attacker identify the remaining individual. 

Lastly, dataset-level defenses largely rely on curation and algorithmic safeguards. A common approach is PII scrubbing, which uses Named Entity Recognition (NER) to detect and remove sensitive spans. In practice, NER systems do not achieve perfect recall across entity types, so substantial PII often remains after scrubbing \cite{pham2025can_llm_recognize,NER_survey}. Moreover, \citet{lukas2023analyzing} demonstrate that attackers can still reconstruct PII present in pre-training corpora even when models are trained on scrubbed data and, in some cases, even under differentially private training; highlighting the limits of current mitigation strategies.

\subsection{Distinction from Conventional Membership Inference Attacks}

Our proposed threat model departs from traditional MIAs in both granularity and objective. Conventional MIAs typically operate at the level of entire samples or prompts. The attacker is given a complete input and aims to infer whether that full instance was present in the model's training dataset. In contrast, our entity-focused threat model is designed to assess entity-level membership, targeting specific entity values (specially PIIs, e.g., names, phone numbers) inserted into redacted templates.

More specifically, rather than asking, ``Was this entire sentence seen during training?'', our attack asks, ``Was this particular entity value used in this specific slot in a training sample?'' For example, given a sentence like ``<full name> works at the Enron Corp.'', the attacker evaluates whether the model's behavior differs when inserting ``John Doe'' versus ``John Smith'', and uses this information to infer whether ``John Doe works at the Enron Corp.'' was part of the training set, leading to violating even user privacy in case of attack success.

This distinction introduces a finer-grained and arguably more realistic privacy threat. In this work we try to highlight that models may exhibit differential behavior even when only one sensitive attribute is altered, underscoring the need for new attack methods as well as defenses that account for token-level or entity-level memorization risk.


In summary, our threat model is complementary to and more targeted than general MIAs, and is specifically motivated by the privacy risks associated with PII exposure in generative language models.

\section{The EL-MIA Benchmark}\label{sec:dataset}

We introduce the \textbf{EL-MIA Benchmark}, a suite comprising the \textit{EL-MIA dataset} and the \textit{set of trained model checkpoints}. Together, these resources provide a controlled setting for evaluating entity-level membership inference focusing on PII, enabling reproducible and systematic analysis across diverse PII types, model scales and training dynamics.


We build our dataset on top of the AI4Privacy~\cite{ai4privacy_2023} open-source dataset, which is designed to resemble natural, real-world text while containing a rich variety of PII. In this corpus, PII spans have already been annotated, covering multiple categories such as first names, last names, phone numbers, email addresses, and other identifying fields. These annotations allow us to directly locate and manipulate PII instances for our membership inference experiments.

Using this base data, we construct our entity-level MIA benchmark \textbf{EL-MIA} by treating each sentence as a template in which all but one PII slots remain fixed, and the remaining slot becomes the target for inference. For each target slot, we generate two sentence variants: \textbf{Member}: the original sentence containing the true PII value from the dataset, and \textbf{Non-member}: the same sentence with the target PII replaced by a randomly selected value of the same type drawn from elsewhere in the corpus. This setup ensures that the two versions differ only in the specific PII under test, enabling clean membership classification experiments. Because PII categories are varied, the dataset supports analysis across heterogeneous types, including structured identifiers (e.g., phone numbers) and free-text fields (e.g., names).

Furthermore, we release three subsets of the datasets based on the distribution of \textit{non-member} values: 
\textbf{Trained}: Non-members are sampled only from the candidates on which the model has been trained on.
\textbf{Untrained}: Non-members are sampled only from the candidates on which the model has \textbf{not} been trained on.
\textbf{Mix}: Non-members are sampled from the both \textit{trained} and \textit{untrained} sets at equal rate.

These subsets can make the analysis of the results easier to apply to the real-world scenario. The \textbf{Untrained} subset is to simulate the easiest for the attackers, because all non-member candidates have not been seen by the model, therefore the model would generally give higher perplexity to the non-members and there would be a larger gap between the member and non-members. Contrarily, \textbf{trained} subset is aimed to simulate the most difficult case, where the members are the closest to non-members. Lastly, \textbf{mix} subset aims to simulate the closest scenario to the real-world, where the attacker obtains some samples (for example using data extraction attacks) which could be memorized entities or hallucinated. 
\vspace{-.1cm}


\section{Methods and Experiments}
In this section, we first benchmark prior MIA methods on the EL-MIA benchmark. After identifying their shortcomings on this task, we introduce novel methods that improve the attack success rate (ASR) over previous approaches.

\subsection{Prior methods}
\label{subsec:prior-benchmarks}

We include four commonly used methods for MIA in our analysis of the \textsc{EL-MIA}. 

\textbf{Lowest loss \cite{carlini2021extracting}.} 
This method scores examples by the tokenized average NLL of the context:
members are expected to have lower loss (higher likelihood).

\textbf{Zlib \cite{carlini2021extracting}.}
This method compares the model's perplexity to the text's zlib compression entropy, using the ratio to measure how ``predictable'' or memorized a sequence is, lower ratios suggest likely membership.

\textbf{Min k\% probs \cite{min_k_prob}.} 
The method checks membership by scoring the sample token-wise, keeping the k\% tokens the model is least confident about, and averaging their log-likelihood. Higher average indicates higher likelihood of membership.

\textbf{Recall \cite{xie-etal-2024-recall}.}
It quantifies the change in conditional log-likelihood induced by prefixing the target example with non-member context. ``Recall'' method achieved the highest ASR on the various MIA benchmarks.

These methods serve as our \textbf{baselines} for evaluating \textsc{EL-MIA}. Since these methods were not designed for entity-level MIA, in the next section, we introduce new attacks to improve the ASR of these baselines.

\subsection{Proposed methods}
\paragraph{Reference-set normalization}

We propose a reference-set normalization attack method that scores how much the sensitive entity candidate $e$ is favored by the model relative to a matched reference set of plausible alternatives ($E_{\text{ref}}=\{e'_1,\dots,e'_N\}$). Given the redacted context $S$, we query the model in a black-box manner to obtain token-level log-likelihoods and compute a conditional score for insertion of each reference element $e'_i$ given $S$. We employ \textit{log-likelihood ratio} or its standardized variant in this decision as follows.

\resizebox{\linewidth}{!}{$
\mathrm{LLR}(e\,|\,S)
= \underbrace{\log P_{\theta}(e\,|\,S)}_{\text{candidate fit}}
  - \underbrace{\log\!\Big(\tfrac{1}{N}\sum_{i=1}^{N} P_{\theta}(e'_i\,|\,S)\Big)}_{\text{reference set}}
$}

In our analysis, for each candidate, we build a type-consistent reference set $E_{ref}$ (same attribute class: name, URL, phone, etc.) matched on simple surface cues with the candidate (token length, script/casing, frequency bin) while excluding the candidate itself. This design is directly practical beyond the benchmark: common entity types are readily synthesizable from public lists and format grammars (e.g., name lexicons, URL/email templates, E.164 phone patterns), enabling black-box deployment without training-data access. We present our experiments considering 5 references for each sample.

\paragraph{Enhancing signal with suffix scoring}

As a simple heuristic to further sharpen the membership signals, we score only the suffix immediately following the candidate span. Concretely, after the candidate boundary, we apply a continuation window of $w$ tokens (ranging from 0 up to the full remaining number of tokens) and compute the token scores within this window, masking out the prefix. This focuses on memorized local continuations (e.g., templated phrasing) while discarding generic context that inflates noise. In all of the figures and tables we refer to this method as ``Reference-set (suffix)''. We also apply this strategy to the Lowest loss baseline, which we refer to as ``Lowest loss (suffix)''.

\subsection{Experimental setup}
\paragraph{Models} We use the Pythia family of models \cite{pythia} of varying scales, ranging from 160M to 6.9B parameters and further pre-train them on our constructed EL-MIA dataset with the causal language modeling objective. Each model is trained for 4 epochs on our dataset, and checkpoints are stored at the end of every epoch. This design allows us to study not only the effect of model capacity on memorization and  vulnerability against MIA, but also how these risks evolve throughout the course of training. By releasing the resulting checkpoints publicly via the Hugging Face model hub, we provide the community with a standard set of baselines that can be directly used to benchmark new attack strategies or defenses on the EL-MIA dataset.\footnote{Trained model checkpoints are accessible at \url{https://huggingface.co/collections/alistvt/el-mia}
} 
In this way, our work contributes both a carefully controlled experimental setting and a set of reproducible artifacts that we hope will accelerate research in auditing and mitigating entity-level memorization in LLMs.


\paragraph{Evaluation}
To evaluate the performance of different methods on the EL-MIA benchmark, we use the commonly used MIA evaluation metrics. We report the area under the attack's ROC curve (AUC) and TPR under low FPR (5\% FPR). Attack's AUC summarizes the trade-off between true positive and false positive rates across varying decision thresholds, providing a threshold-independent indicator of separability between in-training and out-of-training samples. We report the weighted average AUC and TPR at low FPR for different PII categories present in the EL-MIA dataset to gain a better understanding of the performance of the attacks on different PII types.\footnote{
Results of global thresholding is presented Appendix \ref{app:global}
} 







\section{Results}
In this section, we present experimental results guided by six research questions:

\noindent \textbf{RQ1.} How do the prior and our proposed methods perform on the EL-MIA? (\ref{sec:modelcomarison})

\noindent \textbf{RQ2.} Can the performance of the attack be explained by the length of sensitive attribute, prefix, or the whole sample? (\ref{sec:token_count})

\noindent \textbf{RQ3.} Do different sensitive attributes show different susceptibility to EL-MIA methods? (\ref{sec:attribute_type})

\noindent \textbf{RQ4.} What is the effect of the number of entities in a sample? 
(\ref{sec:marginal_info})

\noindent \textbf{RQ5.} How does the model size affect the susceptibility? (\ref{sec:model_size})

\noindent \textbf{RQ6.} How does the entity level susceptibility changes w.r.t the training epochs? (\ref{sec:training_epochs})

\begin{figure*}[ht]
    \centering
    \includegraphics[width=1\linewidth]{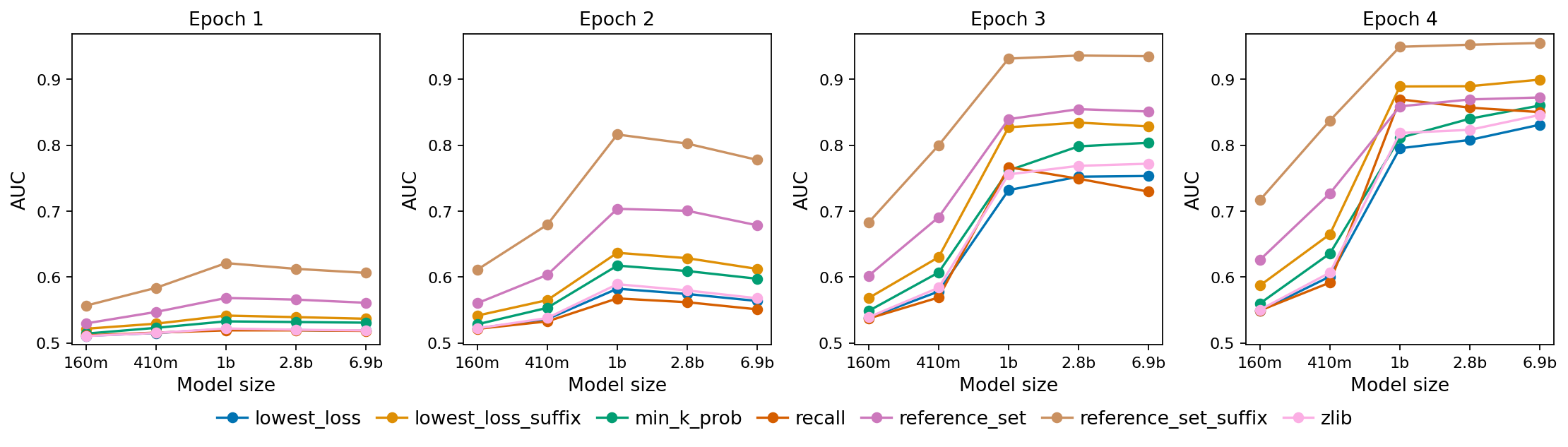}
    \caption{Analysis of all methods in terms of AUC by model size after different epochs of training}
    \label{fig:auc_by_model_size}
\end{figure*}

\subsection{Comparison of the methods} \label{sec:modelcomarison}
The main results are presented in Table \ref{tab:mia_methods_by_model}, comparing seven EL-MIA methods across five model sizes and our three dataset subsets, after 1 epoch of training. Firstly, the proposed reference-set attacks dominate: the reference-set (suffix) variant is consistently best for every model size and subset,  peaking interestingly for the 1B untrained split. Its non-suffix version is second best, and ``lowest loss (suffix)'' also beats its baseline (Lowest loss). Among the baselines, Min-k\% prob generally outperforms others, but still remains bellow the task-specific proposed methods. Across subsets, scores are typically highest on untrained, then mix, and lowest on trained as expected (see Section \ref{sec:dataset}). Overall, these results highlight the limitations of the general MIA methods to the proposed threat model. 

\begin{table*}[t]
\centering
\small
\setlength{\tabcolsep}{6pt}
\begin{tabular}{llcccccc}
\toprule
 & & \multicolumn{2}{c}{\textbf{mix}} & \multicolumn{2}{c}{\textbf{untrained}} & \multicolumn{2}{c}{\textbf{trained}} \\
\textbf{Model Size} & \textbf{Method} & AUC & TPR$_{5\%}$ & AUC & TPR$_{5\%}$ & AUC & TPR$_{5\%}$ \\
\cmidrule(r){3-4}\cmidrule(r){5-6}\cmidrule(l){7-8}
\multirow{7}{*}{160m} & Lowest loss & 51.07 & 5.56 & 51.36 & 5.72 & 50.80 & 5.45 \\
 & Zlib & 51.03 & 5.53 & 51.33 & 5.51 & 50.77 & 5.48 \\
 & Min k\% prob & 51.39 & 5.69 & 51.89 & 5.96 & 50.93 & 5.58 \\
 & Recall & 51.13 & 5.78 & 51.14 & 5.59 & 51.14 & 5.91 \\
& \cellcolor{ourrow} Lowest loss (suffix) & 52.14 & 6.03 & 52.62 & 6.29 & 51.68 & 5.76 \\
& \cellcolor{ourrow} Reference-set & 52.93 & 6.88 & 53.84 & 6.99 & 52.10 & 5.84 \\
& \cellcolor{ourrow} Reference-set (suffix) & \textbf{55.64} & \textbf{7.75} & \textbf{56.81} & \textbf{7.92} & \textbf{54.51} & \textbf{7.65} \\
\midrule
\multirow{7}{*}{410m} & Lowest loss & 51.51 & 5.63 & 51.83 & 5.81 & 51.15 & 5.60 \\
 & Zlib & 51.51 & 5.75 & 51.86 & 5.84 & 51.16 & 5.75 \\
 & Min k\% prob & 52.29 & 6.19 & 52.90 & 6.38 & 51.66 & 5.84 \\
 & Recall & 51.62 & 6.14 & 51.70 & 6.32 & 51.35 & 6.15 \\
& \cellcolor{ourrow} Lowest loss (suffix) & 52.92 & 6.64 & 53.50 & 6.75 & 52.33 & 6.30 \\
& \cellcolor{ourrow} Reference-set & 54.66 & 7.40 & 55.88 & 7.62 & 53.55 & 6.72 \\
& \cellcolor{ourrow} Reference-set (suffix) & \textbf{58.36} & \textbf{8.81} & \textbf{59.60} & \textbf{8.95} & \textbf{57.10} & \textbf{8.22} \\
\midrule
\multirow{7}{*}{1b} & Lowest loss & 52.12 & 5.82 & 52.56 & 6.06 & 51.64 & 5.81 \\
 & Zlib & 52.19 & 6.15 & 52.67 & 6.38 & 51.70 & 6.06 \\
 & Min k\% prob & 53.28 & 6.65 & 54.05 & 7.18 & 52.47 & 6.50 \\
 & Recall & 51.85 & 6.47 & 52.14 & 6.74 & 51.72 & 6.26 \\
& \cellcolor{ourrow} Lowest loss (suffix) & 54.18 & 7.43 & 54.94 & 7.82 & 53.31 & 6.94 \\
& \cellcolor{ourrow} Reference-set & 56.86 & 7.93 & 58.42 & 9.33 & 55.16 & 7.22 \\
& \cellcolor{ourrow} Reference-set (suffix) & \underline{\textbf{62.19}} & \underline{\textbf{11.09}} & \underline{\textbf{63.58}} & \underline{\textbf{11.88}} & \underline{\textbf{60.55}} & \underline{\textbf{10.40}} \\
\midrule
\multirow{7}{*}{2.8b} & Lowest loss & 51.95 & 6.00 & 52.36 & 6.25 & 51.52 & 5.76 \\
 & Zlib & 52.00 & 6.14 & 52.43 & 6.38 & 51.56 & 6.02 \\
 & Min k\% prob & 53.16 & 6.85 & 53.88 & 7.23 & 52.46 & 6.47 \\
 & Recall & 52.05 & 6.61 & 52.03 & 6.64 & 51.63 & 6.64 \\
& \cellcolor{ourrow} Lowest loss (suffix) & 53.94 & 7.15 & 54.64 & 7.56 & 53.15 & 6.82 \\
& \cellcolor{ourrow} Reference-set & 56.51 & 8.14 & 58.14 & 8.96 & 55.06 & 6.91 \\
& \cellcolor{ourrow} Reference-set (suffix) & \textbf{61.18} & \textbf{10.78} & \textbf{62.71} & \textbf{11.30} & \textbf{59.80} & \textbf{10.04} \\
\midrule
\multirow{7}{*}{6.9b} & Lowest loss & 51.84 & 5.71 & 52.21 & 6.04 & 51.46 & 5.61 \\
 & Zlib & 51.88 & 6.13 & 52.29 & 6.28 & 51.49 & 6.01 \\
 & Min k\% prob & 53.07 & 6.64 & 53.75 & 6.99 & 52.38 & 6.29 \\
 & Recall & 51.74 & 6.53 & 52.18 & 6.82 & 51.55 & 6.40 \\
& \cellcolor{ourrow} Lowest loss (suffix) & 53.67 & 7.23 & 54.35 & 7.51 & 52.97 & 7.00 \\
& \cellcolor{ourrow} Reference-set & 56.10 & 7.52 & 57.26 & 8.76 & 54.90 & 6.77 \\
& \cellcolor{ourrow} Reference-set (suffix) & \textbf{60.72} & \textbf{9.92} & \textbf{61.93} & \textbf{10.76} & \textbf{59.22} & \textbf{9.56} \\
\bottomrule
\end{tabular}
\caption{Performance of each attack (rows) on each dataset subset (columns), grouped by model size, for attribute type specific thresholding. Only 1 epoch results are shown. The methods marked in purple are the newly proposed methods. Boldface indicates the best result per model size; underline indicates the best result over all model sizes. }
\label{tab:mia_methods_by_model}
\end{table*}

\subsection{Effect of token count}\label{sec:token_count}

We analyze the TPR based on the sensitive attribute token count, prefix token count (the text that appears earlier than the sensitive attribute), and the whole context token count. Correlation results for the ``reference-set'' attack are summarized in Table \ref{tab:tpr-length-heat}.
We find that prefix tokens length starts contributing positively to the attack success after epoch 1 for all model sizes and larger models generally utilize it more. On the other hand, for the context length and candidate length most of the models do not show any correlations and the two largest models show to be utilizing these only at the 4th epoch.\footnote{Detailed statistics of each of the 56 entity types is presented in Appendix \ref{app:attr}.} 

\begin{table}[t]
\centering
\small
\setlength{\tabcolsep}{6pt}
\renewcommand{\arraystretch}{1.15}
\begin{tabular}{llccc}
\toprule
& & \multicolumn{3}{c}{\textbf{Tokens length}} \\
\textbf{Model} & \textbf{Epoch} & {Prefix} & {Context} & {Candidate} \\
\midrule
\multirow{4}{*}{160m} & 1 & 0.17 & 0.15 & 0.17 \\
 & 2 & \cellcolor[rgb]{1.000,0.718,0.718}0.28 & 0.17 & 0.19 \\
 & 3 & \cellcolor[rgb]{1.000,0.603,0.603}0.40 & 0.18 & 0.23 \\
 & 4 & \cellcolor[rgb]{1.000,0.574,0.574}0.43 & 0.14 & 0.18 \\
 \midrule
\multirow{4}{*}{410m} & 1 & 0.14 & -0.15 & -0.11 \\
 & 2 & \cellcolor[rgb]{1.000,0.712,0.712}0.29 & 0.15 & 0.17 \\
 & 3 & \cellcolor[rgb]{1.000,0.687,0.687}0.31 & 0.05 & 0.09 \\
 & 4 & \cellcolor[rgb]{1.000,0.614,0.614}0.39 & 0.14 & 0.17 \\
 \midrule
\multirow{4}{*}{1b} & 1 & 0.25 & -0.12 & -0.12 \\
 & 2 & \cellcolor[rgb]{1.000,0.647,0.647}0.35 & 0.08 & 0.11 \\
 & 3 & \cellcolor[rgb]{1.000,0.605,0.605}0.39 & 0.14 & 0.15 \\
 & 4 & \cellcolor[rgb]{1.000,0.534,0.534}0.47 & 0.18 & 0.19 \\
 \midrule
\multirow{4}{*}{2.8b} & 1 & 0.14 & 0.02 & 0.10 \\
 & 2 & \cellcolor[rgb]{1.000,0.678,0.678}0.32 & 0.10 & 0.08 \\
 & 3 & \cellcolor[rgb]{1.000,0.571,0.571}0.43 & 0.18 & 0.18 \\
 & 4 & \cellcolor[rgb]{1.000,0.548,0.548}0.45 & \cellcolor[rgb]{1.000,0.735,0.735}0.27 & 0.26 \\
 \midrule
\multirow{4}{*}{6.9b} & 1 & 0.14 & -0.14 & -0.15 \\
 & 2 & \cellcolor[rgb]{1.000,0.724,0.724}0.28 & 0.06 & 0.10 \\
 & 3 & \cellcolor[rgb]{1.000,0.531,0.531}0.47 & 0.21 & 0.20 \\
 & 4 & \cellcolor[rgb]{1.000,0.507,0.507}0.49 & \cellcolor[rgb]{1.000,0.725,0.725}0.28 & \cellcolor[rgb]{1.000,0.736,0.736}0.26 \\
\bottomrule
\end{tabular}
\caption{Heat-colored Pearson's correlations ($r$) between TPR and samples' tokens length metrics for the ``reference-set'' attack. Color scale maps $r=-1$ to blue, $r=0$ to white, $r=1$ to red. Cells that are not statistically significant ($p>0.05$) are left uncolored.}
\label{tab:tpr-length-heat}
\end{table}


\subsection{Effect of sensitive attribute types}\label{sec:attribute_type}
Examining the best attack per entity type, we find that the most vulnerable categories are small, low-entropy, often single-token domains such as SEX (e.g., \{male, female\}), ORDINALDIRECTION (\{N,E,S,W\}), or CURRENCYCODE (\{USD, EUR, ...\}). In such cases, the model can assign sharply peaked probabilities to a handful of options, making membership cues easier to detect and leading to higher AUC/TPR.\footnote{You can refer to Appendix \ref{app:attr} for detailed performance on each of the entity types.}

Motivated by this, we analyze the relation between vulnerability and the number of unique candidates per attribute with a regression analysis. We find a small negative correlation ($r=-0.150$) which suggests that attributes with fewer unique values tend to be more susceptible (higher AUC) to attacks, but the relation is not significant 
($p=0.271$). 

\subsection{Effect of number of entities}\label{sec:marginal_info}

\begin{figure*}[ht]
    \centering
    \includegraphics[width=2.1\columnwidth]{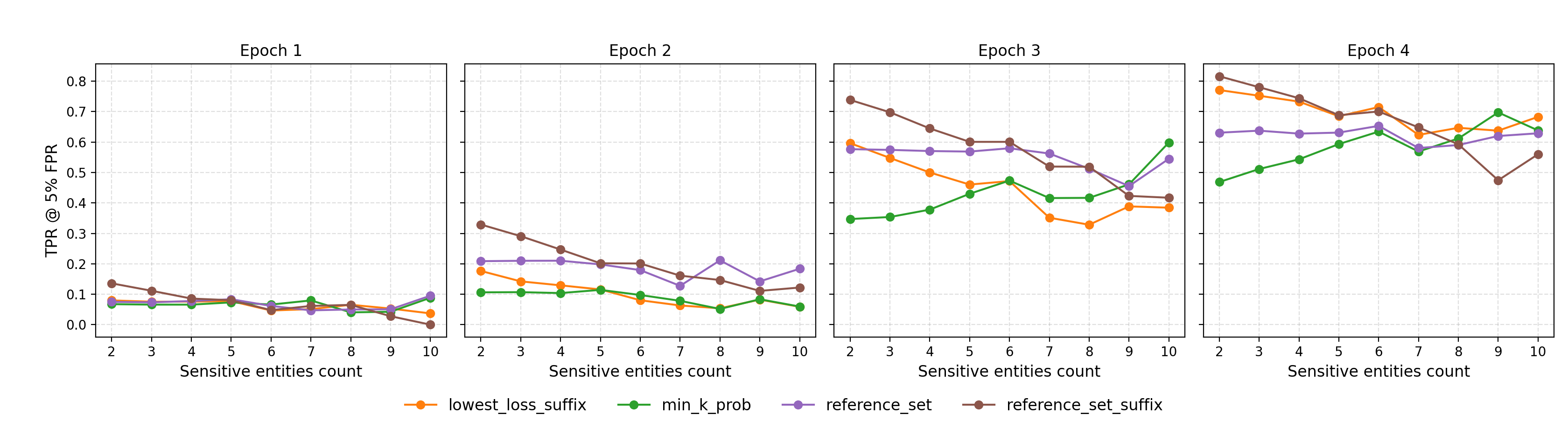}
    
    \caption{Attack TPR at 5\% FPR   explained by number of sensitive entities present in the sample, for the 6.9b model across different epochs. For brevity only best methods are shown.
    }
    \label{fig:marginal_pearson_r}
\end{figure*}

To quantify the relation between the number entities in each sample and MIA vulnerability of the candidate of interest, we group the dataset based on the number of sensitive entities included in each sample, then we report the TPR at 5\% FPR rate of each group.\footnote{Note that the same approach cannot be done for the AUC, as it cannot be measured per sample.} Figure \ref{fig:marginal_pearson_r} shows the plot for the 6.9B model; the results for the other model sizes are similar.  
We find a negative correlation across all methods up to epoch 3, however, for large models (2.8B and 6.9B) after the third epoch the value gets positive. This suggests in the earlier epochs of training, complex samples would benefit from further privacy, however, after more training, large models would remind those samples more easily, as they would start to attend to more complex cues. 

\subsection{Model size}\label{sec:model_size}
When we look at the susceptibility as a function of model size (at fixed epoch) and training dynamics (across epochs) (Figure \ref{fig:auc_by_model_size}), we observe that across epochs, AUC rises for all attacks. With the model size, AUC seems to saturate around 1B, hinting that capacity stops being the bottleneck. This could be due to the fact that larger models can memorize and stabilize entity-specific likelihoods faster, and more epochs amplify this separation. The 1B model at the first 2 epochs remains the most susceptible model which at first is counter-intuitive, however this phenomena is consistent with the bias–variance balance: very large models haven't specialized yet, while mid-size models already overfit to the data \cite{belkin2019reconciling}.



\subsection{Training dynamics}\label{sec:training_epochs}
Susceptibility increases with additional epochs for all sizes, but the rate is size-dependent: larger models show a sharp rise between epochs 2 to 3 and then begin to level off by epoch 4, while smaller models grow more modestly and saturate lower. Taken together (with the model size behavior), these results challenge the earlier findings that bigger models exhibit more memorization \cite{carlini2023quantifying}: while at the final epoch models do scale in susceptibility, early epoch behavior can invert that ranking, with a 1B model peaking first.



\section{Conclusions}
In this paper we formulated a new threat model for auditing the presence of privacy sensitive information in the training data of LLMs: Entity-Level Membership inference attack (EL-MIA). We created a benchmark dataset and evaluated existing MIA methods as well as two newly proposed methods on the proposed dataset.
We found that the existing MIA methods struggle with membership inference on the sensitive entities but our newly proposed methods outperforms all other methods in all model sizes by noticing the deviation of the model likelihood for the entity under interest to a set of crafted entities of the same type. Under properly designed attacks, the LLMs exhibit substantial vulnerability even after 1 epoch of training, underscoring both the validity of our threat model and the remaining room for improvement.

By releasing the benchmark together with resulting model checkpoints publicly via the Hugging Face model hub, we provide the community with a standard set of baselines that can be directly used to benchmark new attack strategies or defenses on the EL-MIA dataset.
In this way, our work contributes both a carefully controlled experimental setting and a set of reproducible artifacts that we hope will accelerate research in auditing and mitigating entity-level memorization in LLMs.



Our paper is also a call for standardized evaluation of PII exposure, which is a realistic threat in operational LLM deployments. Future work should focus on expanding to other access levels and variations of the auxiliary data available to the adversary, the development of further methods, and proposing defense strategies for this threat model.






\section*{Limitations}
  In this work we study MIA arising from further \emph{pre-training} of LLMs. Given evidence that fine-tuning can be misaligned with pre-training~\cite{zeng-etal-2024-exploring}, future work can address this gap. 
    Our threat model assumes access to a verbatim redacted prompt (masked entity, exact template). However, sometimes the attacker may hold paraphrased or approximate templates; constructing a paraphrased EL-MIA benchmark (e.g., LLM-generated variants beyond AI4Privacy) is considered beyond the current work due to the dataset access and replicability, therefore we present the EL-MIA benchmark as a first step and the approximate MIA on the entities remains a key but challenging direction for future work.
  Lastly, our dataset and results are limited to English. Given cross-lingual differences in language structure and privacy dynamics \cite{xlm_privacy}, future work should evaluate additional languages to test whether the observed patterns persist or diverge.


\section*{Acknowledgements}
This publication is part of the project LESSEN\footnote{\url{https://lessen-project.nl}} with project number NWA.1389.20.183 of the research program NWA ORC 2020/21 which is (partly) financed by the Dutch Research Council (NWO).

We thank the Center for Information Technology of the University of Groningen for their support and for providing access to the Hábrók high performance computing cluster.

This work used the Dutch national e-infrastructure with the support of the SURF Cooperative using grant no. EINF-13489.
\bibliography{custom}
\bibliographystyle{plainnat}

\clearpage            
\onecolumn            

\appendix

\clearpage

\section{Global thresholding}\label{app:global}
Results of the attacks when global thresholding (opposed to entity-type-specific thresholding) is used are presented in Table \ref{tab:global_thresh}. All of the results in this table are slightly lower than their respective cell in the entity-type-specific thresholding results (Table \ref{tab:mia_methods_by_model}).

\begin{table*}[t]
\centering
\small
\setlength{\tabcolsep}{6pt}
\begin{tabular}{llcccccc}
\toprule
 & & \multicolumn{2}{c}{\textbf{mix}} & \multicolumn{2}{c}{\textbf{untrained}} & \multicolumn{2}{c}{\textbf{trained}} \\
\textbf{Model Size} & \textbf{Method} & AUC & TPR$_{5\%}$ & AUC & TPR$_{5\%}$ & AUC & TPR$_{5\%}$ \\
\cmidrule(r){3-4}\cmidrule(r){5-6}\cmidrule(l){7-8}
\multirow{7}{*}{160m} & Lowest loss & 50.94 & 5.41 & 51.16 & 5.49 & 50.74 & 5.32 \\
 & Zlib & 50.94 & 5.44 & 51.19 & 5.46 & 50.74 & 5.42 \\
 & Min k\% prob & 51.28 & 5.54 & 51.73 & 5.59 & 50.88 & 5.59 \\
 & Recall & 51.09 & 5.70 & 51.11 & 5.66 & 51.09 & 5.87 \\
& \cellcolor{ourrow} Lowest loss (suffix) & 51.68 & 5.67 & 51.95 & 5.84 & 51.37 & 5.48 \\
& \cellcolor{ourrow} Reference-set & 52.63 & 6.10 & 53.38 & 6.17 & 51.90 & 5.80 \\
& \cellcolor{ourrow} Reference-set (suffix) & 55.49 & 7.17 & 56.65 & 7.46 & 54.14 & 7.27 \\
\midrule
\multirow{7}{*}{410m} & Lowest loss & 51.32 & 5.48 & 51.58 & 5.56 & 51.05 & 5.42 \\
 & Zlib & 51.36 & 5.42 & 51.64 & 5.45 & 51.07 & 5.30 \\
 & Min k\% prob & 52.11 & 5.78 & 52.68 & 5.82 & 51.54 & 5.74 \\
 & Recall & 51.49 & 5.85 & 51.58 & 6.18 & 51.29 & 6.05 \\
& \cellcolor{ourrow} Lowest loss (suffix) & 52.24 & 5.91 & 52.59 & 6.00 & 51.86 & 5.67 \\
& \cellcolor{ourrow} Reference-set & 53.93 & 6.56 & 54.97 & 6.51 & 53.03 & 6.21 \\
& \cellcolor{ourrow} Reference-set (suffix) & 57.92 & 8.09 & 59.27 & 8.34 & 56.62 & 7.50 \\
\midrule
\multirow{7}{*}{1b} & Lowest loss & 51.84 & 5.54 & 52.19 & 5.78 & 51.48 & 5.40 \\
 & Zlib & 51.94 & 5.82 & 52.32 & 5.97 & 51.54 & 5.74 \\
 & Min k\% prob & 53.03 & 6.26 & 53.70 & 6.36 & 52.33 & 6.18 \\
 & Recall & 51.76 & 6.27 & 51.95 & 6.48 & 51.63 & 6.17 \\
& \cellcolor{ourrow} Lowest loss (suffix) & 53.22 & 6.41 & 53.66 & 6.60 & 52.63 & 6.14 \\
& \cellcolor{ourrow} Reference-set & 56.01 & 6.60 & 57.19 & 7.58 & 54.34 & 6.57 \\
& \cellcolor{ourrow} Reference-set (suffix) & 61.56 & 9.96 & 63.13 & 11.26 & 59.76 & 9.04 \\
\midrule
\multirow{7}{*}{2.8b} & Lowest loss & 51.73 & 5.72 & 52.05 & 5.76 & 51.40 & 5.48 \\
 & Zlib & 51.79 & 5.72 & 52.14 & 5.80 & 51.44 & 5.66 \\
 & Min k\% prob & 52.92 & 6.27 & 53.55 & 6.32 & 52.33 & 6.19 \\
 & Recall & 51.92 & 6.51 & 51.87 & 6.40 & 51.58 & 6.23 \\
& \cellcolor{ourrow} Lowest loss (suffix) & 53.06 & 6.52 & 53.48 & 6.60 & 52.52 & 6.23 \\
& \cellcolor{ourrow} Reference-set & 55.79 & 6.85 & 57.22 & 7.32 & 54.54 & 6.58 \\
& \cellcolor{ourrow} Reference-set (suffix) & 60.66 & 9.64 & 62.31 & 10.78 & 59.15 & 9.01 \\
\midrule
\multirow{7}{*}{6.9b} & Lowest loss & 51.63 & 5.60 & 51.94 & 5.68 & 51.33 & 5.52 \\
 & Zlib & 51.70 & 5.80 & 52.04 & 5.88 & 51.39 & 5.70 \\
 & Min k\% prob & 52.86 & 6.32 & 53.47 & 6.42 & 52.25 & 6.24 \\
 & Recall & 51.63 & 6.23 & 52.03 & 6.22 & 51.47 & 6.32 \\
& \cellcolor{ourrow} Lowest loss (suffix) & 52.87 & 6.33 & 53.28 & 6.44 & 52.40 & 6.19 \\
& \cellcolor{ourrow} Reference-set & 55.28 & 6.83 & 56.35 & 7.34 & 54.30 & 6.10 \\
& \cellcolor{ourrow} Reference-set (suffix) & 60.15 & 9.34 & 61.58 & 10.10 & 58.55 & 8.44 \\
\bottomrule
\end{tabular}
\caption{AUC and TPR at 5\% FPR for each method (rows) and dataset subset (columns), grouped by model size, when the thresholds are selected over the whole dataset (rather than attribute-type-specific thresholding). Only 1 epoch results shown. The methods marked in purple are the newly proposed methods.}
\label{tab:global_thresh}
\end{table*}

\clearpage
\section{Effect of number of entities plots}
\begin{figure*}
        \centering
        \includegraphics[width=1\linewidth]{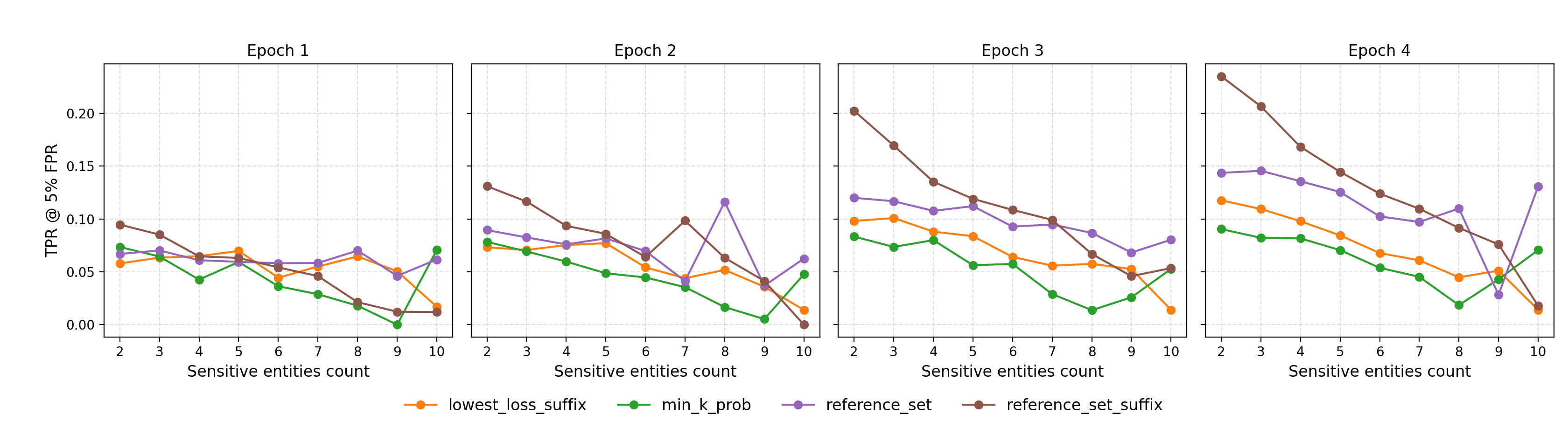}
        \caption{Attack TPR at 5\% FPR   explained by number of sensitive entities present in the sample, for the  160M model across different epochs. For brevity only best methods are shown.}
        \label{fig:marginal_160m}
\end{figure*}

\begin{figure*}
        \centering
        \includegraphics[width=1\linewidth]{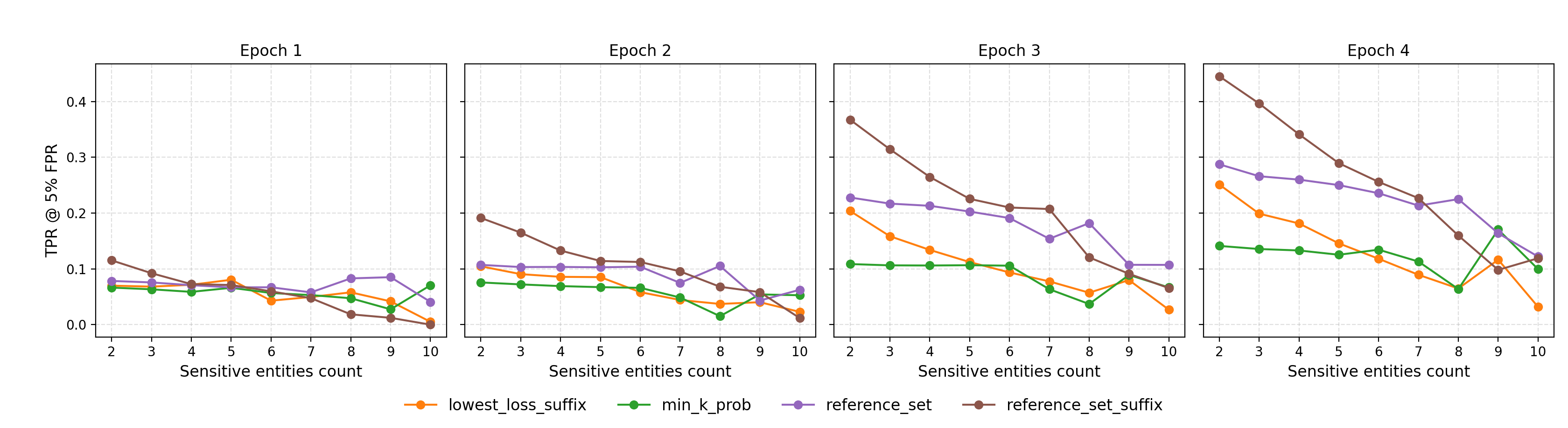}
        \caption{Attack TPR at 5\% FPR   explained by number of sensitive entities present in the sample, for the  410M model across different epochs. For brevity only best methods are shown.}
        \label{fig:marginal_410m}
\end{figure*}

\begin{figure*}
        \centering
        \includegraphics[width=1\linewidth]{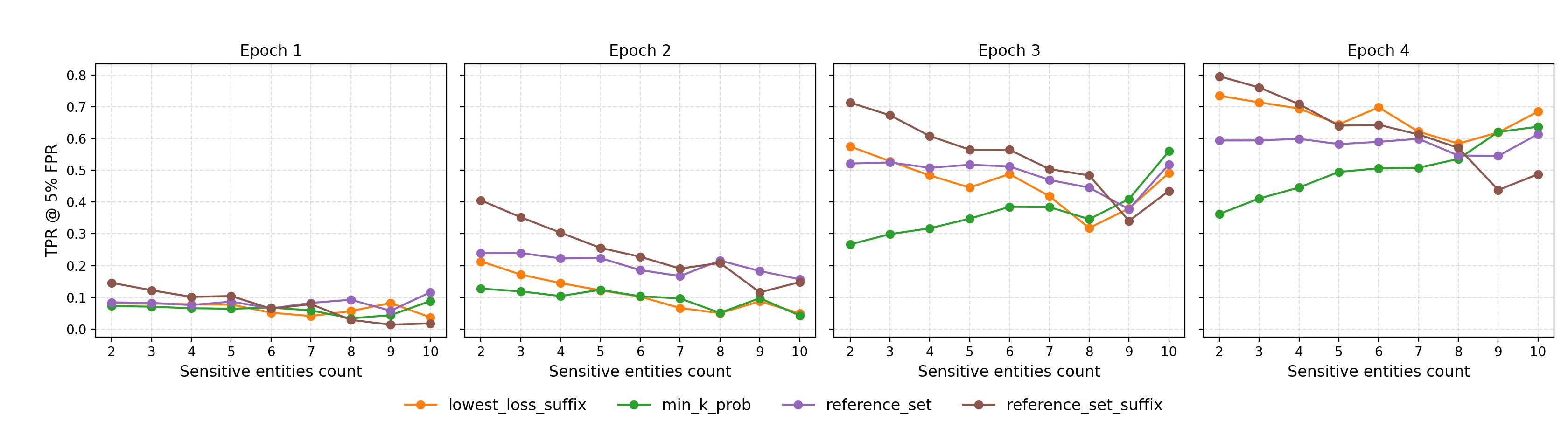}
        \caption{Attack TPR at 5\% FPR   explained by number of sensitive entities present in the sample, for the  1B model across different epochs. For brevity only best methods are shown.}
        \label{fig:marginal_1b}
\end{figure*}

\begin{figure*}
        \centering
        \includegraphics[width=1\linewidth]{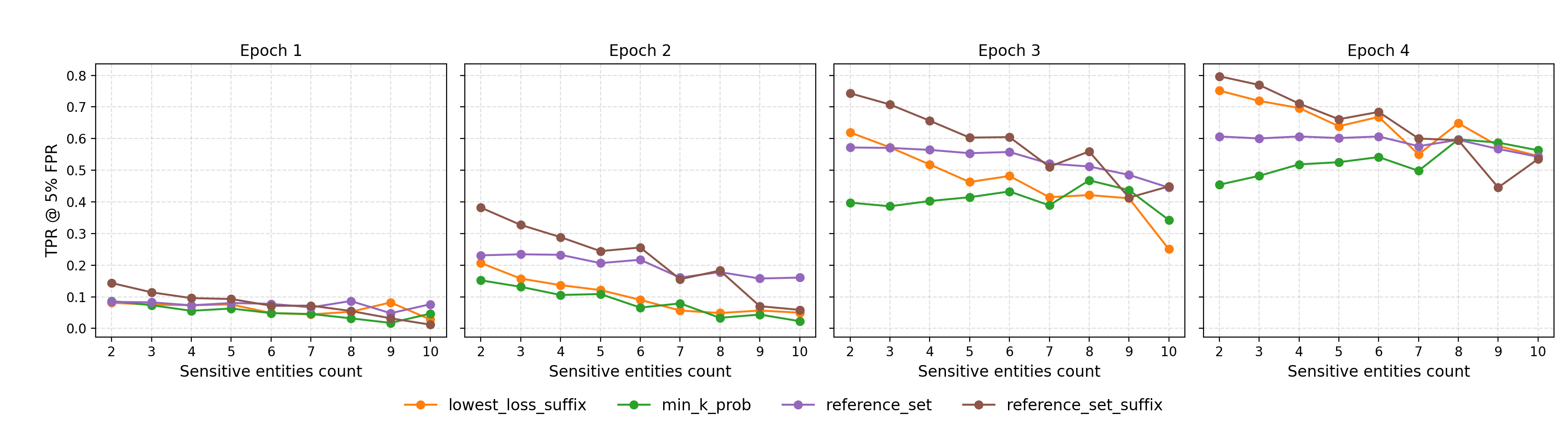}
        \caption{Attack TPR at 5\% FPR   explained by number of sensitive entities present in the sample, for the  2.8B model across different epochs. For brevity only best methods are shown.}
        \label{fig:marginal_2.8b}
\end{figure*}

\begin{figure*}
        \centering
        \includegraphics[width=1\linewidth]{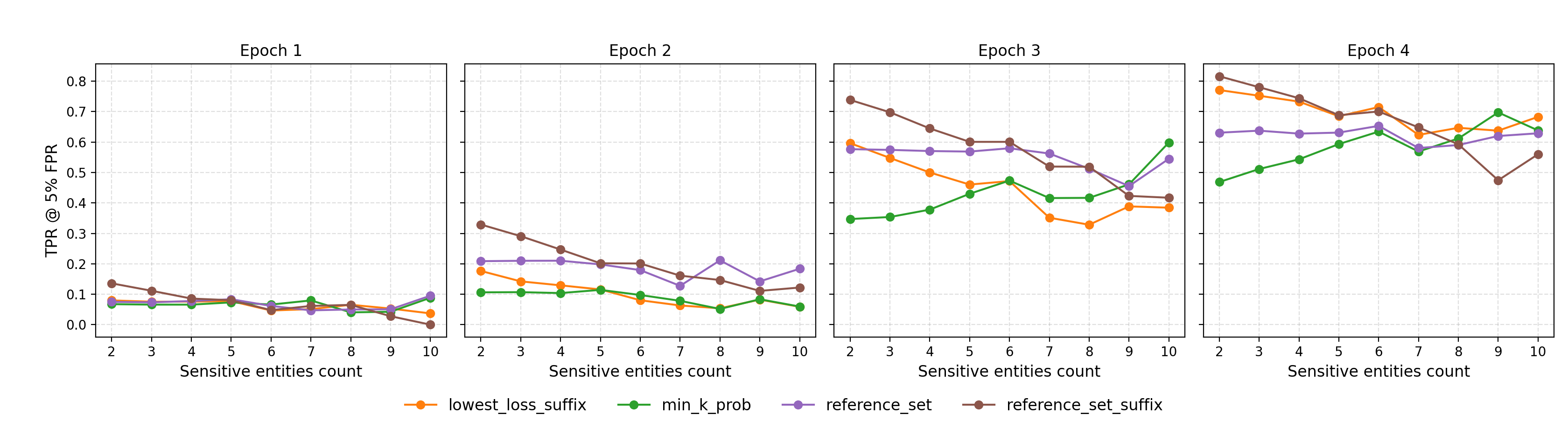}
        \caption{Attack TPR at 5\% FPR   explained by number of sensitive entities present in the sample, for the  6.9B model across different epochs. For brevity only best methods are shown.}
        \label{fig:marginal_6.9b}
\end{figure*}

\begin{figure*}
    \centering
        \includegraphics[width=1\linewidth]{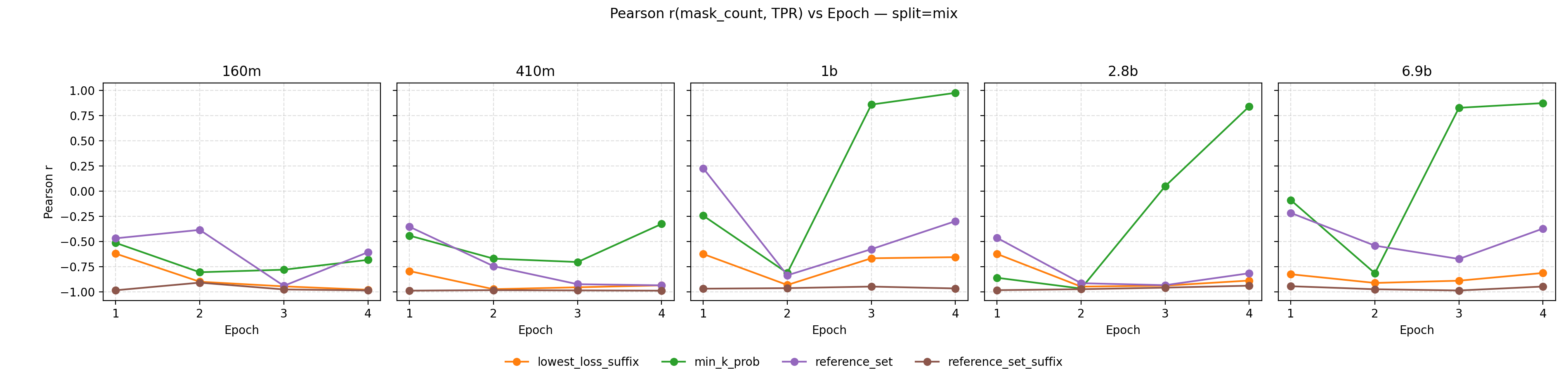}
        \caption{Pearson's correlation between attack TPR @ 5\% FPR and the number of sensitive entities at each sample across different epochs and model sizes. (Derived from Figures \ref{fig:marginal_160m} to \ref{fig:marginal_6.9b})}
        \label{fig:marginal_info_pearson}
\end{figure*}

Attack performance for the ``mix'' split, based on the number of sensitive entities in each sample is presented in Figures \ref{fig:marginal_160m} to \ref{fig:marginal_6.9b}. We further process these results by calculating the Pearson's correlation (Figure \ref{fig:marginal_info_pearson}). As one can see in Figure \ref{fig:marginal_info_pearson}, all methods obtain negative correlation, showing their performance degrades when more sensitive entities are present in the sentence. However, it seems that for larger models (1B onwards), and after then second epoch of training, only the Min K\% Probs method is somehow utilizing the presence of more sensitive attributes in the sample. This trend also is visible in Figures \ref{fig:marginal_1b} to \ref{fig:marginal_6.9b} for the Min K\% Probs method with a slightly positive slope of its curve.


\definecolor{Teal}{RGB}{0,128,128}
\begin{table*}[ht]
\centering
\small
\begin{tabular}{l ll ll lll}
\toprule
& \multicolumn{2}{c}{AUC (\%)} & \multicolumn{2}{c}{TPR (\%)} & \multicolumn{3}{c}{Token Len Median} \\
\cmidrule(lr){2-3}\cmidrule(lr){4-5}\cmidrule(lr){6-8}
Attribute & Method & Value & Method & Value & Candidate & Prefix & Context \\
\midrule
HEIGHT & \textcolor{Black}{reference\_set\_suffix} & 69.95 & \textcolor{Black}{reference\_set\_suffix} & 16.26 & 4 & 19 & 38 \\
CITY & \textcolor{Black}{reference\_set\_suffix} & 69.12 & \textcolor{Black}{reference\_set\_suffix} & 12.56 & 3 & 16 & 40 \\
CURRENCYCODE & \textcolor{Black}{reference\_set\_suffix} & 68.51 & \textcolor{Black}{reference\_set\_suffix} & 14.85 & 2 & 20 & 43 \\
BUILDINGNUMBER & \textcolor{Black}{reference\_set\_suffix} & 67.66 & \textcolor{Violet}{lowest\_loss\_suffix} & 7.02 & 2 & 16 & 37 \\
CREDITCARDCVV & \textcolor{Black}{reference\_set\_suffix} & 65.85 & \textcolor{Black}{reference\_set\_suffix} & 17.07 & 1 & 31 & 39 \\
ZIPCODE & \textcolor{Black}{reference\_set\_suffix} & 65.64 & \textcolor{Black}{reference\_set\_suffix} & 12.54 & 5 & 20 & 38 \\
VEHICLEVRM & \textcolor{Black}{reference\_set\_suffix} & 65.62 & \textcolor{Black}{reference\_set\_suffix} & 18.25 & 4 & 21 & 43 \\
URL & \textcolor{Gray}{reference\_set} & 65.42 & \textcolor{Black}{reference\_set\_suffix} & 14.61 & 10 & 26 & 47 \\
MIDDLENAME & \textcolor{Black}{reference\_set\_suffix} & 65.16 & \textcolor{Black}{reference\_set\_suffix} & 12.41 & 2 & 5 & 41 \\
JOBAREA & \textcolor{Black}{reference\_set\_suffix} & 64.99 & \textcolor{Black}{reference\_set\_suffix} & 10.71 & 2 & 15 & 40 \\
IPV4 & \textcolor{Gray}{reference\_set} & 64.89 & \textcolor{Gray}{reference\_set} & 20.26 & 7 & 23 & 45 \\
SEX & \textcolor{Black}{reference\_set\_suffix} & 64.88 & \textcolor{Black}{reference\_set\_suffix} & 13.86 & 1 & 8 & 37 \\
BIC & \textcolor{Gray}{reference\_set} & 64.48 & \textcolor{Cyan}{min\_k\_prob} & 14.17 & 6 & 27 & 43 \\
FIRSTNAME & \textcolor{Black}{reference\_set\_suffix} & 64.41 & \textcolor{Black}{reference\_set\_suffix} & 12.03 & 2 & 2 & 40 \\
CURRENCYNAME & \textcolor{Black}{reference\_set\_suffix} & 64.04 & \textcolor{Black}{reference\_set\_suffix} & 21.55 & 4 & 21 & 45 \\
IBAN & \textcolor{Gray}{reference\_set} & 64.02 & \textcolor{Black}{reference\_set\_suffix} & 11.94 & 11 & 23 & 46 \\
USERNAME & \textcolor{Black}{reference\_set\_suffix} & 63.87 & \textcolor{Black}{reference\_set\_suffix} & 12.88 & 5 & 10 & 42 \\
STREET & \textcolor{Black}{reference\_set\_suffix} & 63.64 & \textcolor{Black}{reference\_set\_suffix} & 9.47 & 3 & 16 & 39 \\
ORDINALDIRECTION & \textcolor{Black}{reference\_set\_suffix} & 63.53 & \textcolor{Black}{reference\_set\_suffix} & 14.46 & 2 & 17 & 35 \\
MASKEDNUMBER & \textcolor{Black}{reference\_set\_suffix} & 63.43 & \textcolor{Black}{reference\_set\_suffix} & 12.38 & 7 & 21 & 41 \\
STATE & \textcolor{Black}{reference\_set\_suffix} & 62.99 & \textcolor{Black}{reference\_set\_suffix} & 10.40 & 3 & 18 & 41 \\
CURRENCYSYMBOL & \textcolor{Black}{reference\_set\_suffix} & 62.92 & \textcolor{Gray}{reference\_set} & 8.65 & 2 & 19 & 42 \\
PIN & \textcolor{Black}{reference\_set\_suffix} & 62.85 & \textcolor{Black}{reference\_set\_suffix} & 7.83 & 2 & 29 & 40 \\
EMAIL & \textcolor{Gray}{reference\_set} & 62.78 & \textcolor{Violet}{lowest\_loss\_suffix} & 11.97 & 10 & 31 & 46 \\
JOBTYPE & \textcolor{Black}{reference\_set\_suffix} & 62.55 & \textcolor{Black}{reference\_set\_suffix} & 10.36 & 2 & 11 & 37 \\
PHONEIMEI & \textcolor{Gray}{reference\_set} & 62.41 & \textcolor{Black}{reference\_set\_suffix} & 12.88 & 10 & 27 & 47 \\
CURRENCY & \textcolor{Black}{reference\_set\_suffix} & 62.15 & \textcolor{Black}{reference\_set\_suffix} & 15.46 & 4 & 21 & 44 \\
NEARBYGPSCOORDINATE & \textcolor{Black}{reference\_set\_suffix} & 61.49 & \textcolor{Gray}{reference\_set} & 9.45 & 11 & 19 & 44 \\
LITECOINADDRESS & \textcolor{Black}{reference\_set\_suffix} & 61.44 & \textcolor{Cyan}{min\_k\_prob} & 8.26 & 22 & 32 & 72 \\
VEHICLEVIN & \textcolor{Gray}{reference\_set} & 61.15 & \textcolor{black}{recall} & 10.61 & 10 & 21 & 44 \\
GENDER & \textcolor{Black}{reference\_set\_suffix} & 60.57 & \textcolor{Black}{reference\_set\_suffix} & 12.10 & 3 & 7 & 39 \\
ACCOUNTNAME & \textcolor{Black}{reference\_set\_suffix} & 60.52 & \textcolor{Gray}{reference\_set} & 12.02 & 3 & 18 & 40 \\
AGE & \textcolor{Black}{reference\_set\_suffix} & 60.35 & \textcolor{Black}{reference\_set\_suffix} & 10.14 & 2 & 15 & 36 \\
AMOUNT & \textcolor{Gray}{reference\_set} & 60.01 & \textcolor{Gray}{reference\_set} & 13.18 & 4 & 19 & 46 \\
COUNTY & \textcolor{Black}{reference\_set\_suffix} & 59.96 & \textcolor{Black}{reference\_set\_suffix} & 6.86 & 3 & 15 & 38 \\
LASTNAME & \textcolor{Black}{reference\_set\_suffix} & 59.75 & \textcolor{Black}{reference\_set\_suffix} & 9.62 & 3 & 5 & 44 \\
COMPANYNAME & \textcolor{Black}{reference\_set\_suffix} & 59.63 & \textcolor{Violet}{lowest\_loss\_suffix} & 8.45 & 6 & 13 & 46 \\
EYECOLOR & \textcolor{Black}{reference\_set\_suffix} & 58.67 & \textcolor{Black}{reference\_set\_suffix} & 12.03 & 2 & 22 & 38 \\
IP & \textcolor{Gray}{reference\_set} & 58.64 & \textcolor{Gray}{reference\_set} & 9.85 & 7 & 21 & 58 \\
DOB & \textcolor{Black}{reference\_set\_suffix} & 58.46 & \textcolor{Black}{reference\_set\_suffix} & 7.42 & 5 & 17 & 45 \\
SECONDARYADDRESS & \textcolor{Black}{reference\_set\_suffix} & 58.43 & \textcolor{Violet}{lowest\_loss\_suffix} & 10.15 & 3 & 19 & 37 \\
ACCOUNTNUMBER & \textcolor{Black}{reference\_set\_suffix} & 58.43 & \textcolor{black}{recall} & 8.14 & 3 & 24 & 42 \\
PHONENUMBER & \textcolor{Black}{reference\_set\_suffix} & 57.73 & \textcolor{Black}{reference\_set\_suffix} & 10.70 & 7 & 34 & 47 \\
IPV6 & \textcolor{Black}{reference\_set\_suffix} & 57.61 & \textcolor{Cyan}{min\_k\_prob} & 11.08 & 29 & 26 & 69 \\
PREFIX & \textcolor{Black}{reference\_set\_suffix} & 57.56 & \textcolor{Black}{reference\_set\_suffix} & 9.20 & 2 & 2 & 44 \\
CREDITCARDNUMBER & \textcolor{Black}{reference\_set\_suffix} & 57.38 & \textcolor{Gray}{reference\_set} & 10.31 & 7 & 24 & 40 \\
BITCOINADDRESS & \textcolor{Black}{reference\_set\_suffix} & 57.10 & \textcolor{Cyan}{min\_k\_prob} & 10.11 & 24 & 23 & 66 \\
PASSWORD & \textcolor{Black}{reference\_set\_suffix} & 56.64 & \textcolor{black}{recall} & 8.31 & 9 & 33 & 51 \\
JOBTITLE & \textcolor{Black}{reference\_set\_suffix} & 55.98 & \textcolor{black}{recall} & 8.17 & 4 & 13 & 46 \\
MAC & \textcolor{Gray}{reference\_set} & 55.52 & \textcolor{black}{recall} & 7.75 & 14 & 26 & 52 \\
SSN & \textcolor{Violet}{lowest\_loss\_suffix} & 54.34 & \textcolor{Cyan}{min\_k\_prob} & 7.86 & 5 & 21 & 42 \\
CREDITCARDISSUER & \textcolor{Gray}{reference\_set} & 54.32 & \textcolor{Black}{reference\_set\_suffix} & 7.11 & 2 & 25 & 40 \\
DATE & \textcolor{Black}{reference\_set\_suffix} & 53.63 & \textcolor{Violet}{lowest\_loss\_suffix} & 6.49 & 5 & 20 & 44 \\
TIME & \textcolor{Gray}{reference\_set} & 53.33 & \textcolor{Gray}{reference\_set} & 10.78 & 3 & 23 & 42 \\
ETHEREUMADDRESS & \textcolor{black}{recall} & 52.24 & \textcolor{black}{recall} & 7.55 & 27 & 32 & 72 \\
USERAGENT & \textcolor{Gray}{reference\_set} & 52.12 & \textcolor{Black}{reference\_set\_suffix} & 6.90 & 43 & 24 & 81 \\
\bottomrule
\end{tabular}
\caption{Attribute Difficulty Table of 6.9B on the Mix dataset after 1 epochs training. }
\label{tab:attribute_performance}
\end{table*}

\section{Attribute-specific performance}\label{app:attr}
Table \ref{tab:attribute_performance} shows the attribute-specific ASR and the statistics of different attributes in terms of their median token length.

\end{document}